\pdfoutput=1

\documentclass[11pt]{article}

\usepackage{ACL2023}

\usepackage{times}
\usepackage{latexsym}
\usepackage[T1]{fontenc}
\usepackage{float}

\usepackage[utf8]{inputenc}

\usepackage{microtype}
\usepackage{inconsolata}
\usepackage{textcomp}
\usepackage{CJKutf8}
\usepackage{multirow}
\usepackage{graphicx}
\usepackage{comment}
\usepackage{amssymb}
\usepackage{amsmath}
\usepackage{bm}
\title{Rakuten Data Release: \\ A Large-Scale and Long-Term Reviews Corpus for Hotel Domain}

\author{Yuki Nakayama$^\clubsuit$\thanks{\hspace{0.2cm}Corresponding Author} \hspace{1.0cm} Koki Hikichi$^\spadesuit$ \hspace{1.0cm} Yun Ching Liu$^\clubsuit$ \hspace{1.0cm}Yu Hirate$^\clubsuit$\\
\texttt{\{yuki.b.nakayama, koki.hikichi, yunching.liu, yu.hirate\}@rakuten.com}\\
  $^\clubsuit$Rakuten Institute of Technology, Rakuten Group, Inc., Tokyo, Japan\\
  $^\spadesuit$Travel \& Mobility Business, Rakuten Group, Inc., Tokyo, Japan
  }

\begin{document}
\maketitle
\begin{abstract}
This paper presents a large-scale corpus of Rakuten Travel Reviews. Our collection contains 7.29 million customer reviews for 16 years, ranging from 2009 to 2024.
Each record in the dataset contains the review text, its response from an accommodation, an anonymized reviewer ID, review date, accommodation ID, plan ID, plan title, room type, room name, purpose, accompanying group, and user ratings from six aspect categories, as well as an overall score.
We present statistical information about our corpus and provide insights into factors driving data drift between 2019 and 2024 using statistical approaches.

\end{abstract}

\section{Introduction}
\label{sec:intro}
Our data repository, ``Rakuten Data Release''\footnote{\url{https://rit.rakuten.com/data_release/}} (RDR), has enabled academic researchers to undertake data-driven research leveraging consumer data collected across a range of Rakuten services, specifically Rakuten Ichiba\footnote{\url{https://www.rakuten.co.jp}}, Rakuten Travel\footnote{\url{https://travel.rakuten.co.jp}}, Rakuten GORA\footnote{\url{https://gora.golf.rakuten.co.jp}}, and Rakuten Recipe\footnote{\url{https://recipe.rakuten.co.jp}}.
In particular, use of Rakuten Service’s customer reviews has brought remarkable success on recommender systems~\cite{Suzumura:25} and various Natural Language Processing tasks such as word replacement~\cite{Hiraoka:22}, tokenization~\cite{hiraoka:21}, review response generation~\cite{shirai:24}, text normalization~\cite{higashiyama:25}, and sentiment analysis~\cite{Nakayama:15, toyama:17, Nakayama:22}.

However, there is a potential issue of using the current review data in a hotel domain\footnote{Hereafter, we focus only on Rakuten Travel data. If you wish to see data specification for the other services, please access \url{https://rit.rakuten.com/data_release/}}.
\paragraph{Model Robustness and Practicality}
All publicly available reviews in the RDR were posted before 2020. If a model is trained solely on old data, it might overfit to the specific expressions and trends of that era. As a result, the model's performance could degrade when applied to current data that includes recent trends and new features. When operating a model in real-world business, a model evaluated on an outdated test set may fail to accurately analyze current customers' sentiment, thereby diminishing its practical value.

Nowadays, this issue is crucial, especially in the hotel domain.
For instance, the COVID-19 pandemic had a devastating impact on the hospitality industry, fundamentally changing aspects such as business reconstruction, hygiene management, and accelerating the adoption of contactless services in accommodation facilities. Furthermore, with the growth of environmental consciousness, notably driven by the SDGs (Sustainable Development Goals), hotel hospitality practices have been dramatically changed, particularly in terms of cost reduction, such as changes in amenity distribution methods.
To ensure the robustness and practical utility of the model, it is essential to prepare training and test datasets that reflect the current hotel situation of post-COVID-19.

Intending to advance and facilitate research using customer review data in the field of Japanese NLP research and the other related fields from the above perspectives, in this paper, we provide an up-to-date Rakuten Travel reviews corpus as shown in Table~\ref{tb:corpus}, ranging from the pre-COVID era to the post-COVID era. 
Our main contributions are summarized as follows:
(1) To the best of our knowledge, in the hotel domain, no attempt has been made to release a publicly available dataset in which the collection duration is more than 15 years; (2) We distribute our corpus for easy access by any research institute, particularly, university laboratories for academic research purposes, which can be obtained via the IDR repository\footnote{\url{https://www.nii.ac.jp/dsc/idr/rakuten/}}; (3) We give an insight into factors of data drift in hotel review texts with statistical approaches.
\begin{table*}[ht]
\small
 {\tabcolsep=1mm
 \begin{tabular}{ll}
 \hline
  \multicolumn{1}{c}{Column Name}&\multicolumn{1}{c}{Sample Value}\\
  \hline \hline
  Reviewer ID &user\_21\\
  Review Date &2018-05-08 21:15:00\\
  Accommodation ID & 5547\\
  Plan ID & 4174514\\
  \hline
  Plan Title &
  \begin{tabular}{l}
  \begin{CJK}{UTF8}{ipxm}カード決済限定★GWの1室売プラン！4～5名様1室利用がお得！1室60000円税別から\end{CJK}\\
  \begin{CJK}{UTF8}{ipxm}【本館】S21\end{CJK}\\
  (Card Payment Only! Golden Week Room Sale Plan! Great Value for 4-5 Guests per Room! \\ From ¥60,000 (excluding tax) per room $[$Main Building$]$ S21)
  \end{tabular}\\
  \hline
  Room Type&heya5uriy\\
  \hline
  Room Name&
  \begin{tabular}{l}
  \begin{CJK}{UTF8}{ipxm}■5名1室販売■【禁煙】本館和洋室（山側）\end{CJK}\\
  (Room for 5 Guests Sale $[$Non-smoking$]$ Main Building
  Japanese/Western-style Room\\ (Mountain View))
  \end{tabular}\\
  \hline
  Purpose&
 \begin{CJK}{UTF8}{ipxm}レジャー (leisure)\end{CJK} \\
  \hline
  Accompanying Group&\begin{CJK}{UTF8}{ipxm}恋人\end{CJK} (couple)\\
  \hline
  User Rating (Location)&5\\
  User Rating (Room)&5\\
  User Rating (Food)&5\\
  User Rating (Bath)&5\\
  User Rating (Services)&5\\
  User Rating (Amenity)&5\\
  User Rating (Overall)&5\\
  \hline
  Review Body&
  \begin{tabular}{l}
  \begin{CJK}{UTF8}{ipxm}風呂、食事、プール、すべて大満足でした。子供たちも大喜びでした。\end{CJK}
  \\
  \begin{CJK}{UTF8}{ipxm}また利用したいです。\end{CJK}\\

  (The baths, meals, and pool were all extremely satisfying. The children were also delighted. \\We would love to stay again.)
  \end{tabular}\\
  \hline
  \begin{tabular}{l}
  Reply from \\the accomodation
  \end{tabular}
  &
  
  \begin{tabular}{l}
  \begin{CJK}{UTF8}{ipxm}この度は杉乃井ホテルをご利用いただきまして、誠にありがとうございます。\end{CJK}
  \\
  \begin{CJK}{UTF8}{ipxm}お風呂、お食事、プールと全てにご満足いただけたとの事、何よりでございます。\end{CJK}
  \\
  \begin{CJK}{UTF8}{ipxm}お子様達もお喜びになられたとの事、私どもも嬉しい次第でございます。\end{CJK}
  \\
  \begin{CJK}{UTF8}{ipxm}今後も多くのお客様にご満足いただけるホテルでありますよう、\end{CJK}\\
  \begin{CJK}{UTF8}{ipxm}最善を尽くしてまいります。またのお越しを心よりお待ち申し上げております。
  \end{CJK}\\
  \begin{CJK}{UTF8}{ipxm}
  ありがとうございました。
  \end{CJK}
  \\
  (Thank you very much for choosing Suginoi Hotel on this occasion. We are most delighted to \\
  hear that you were completely satisfied with everything, including the baths, meals, and the pool. \\We are also very happy to know that your children enjoyed themselves. We will continue to do our\\ utmost to be a hotel that satisfies many customers in the future. We sincerely look forward to your\\ next visit. Thank you.)
  \end{tabular}\\
  \hline
 \end{tabular}
 }
 \centering
 (a) User Evaluation Data
 \small
 \vspace{0.5cm}

 \begin{tabular}{ll}
\hline
\multicolumn{1}{c}{Column Name} & \multicolumn{1}{c}{Sample Value}\\
\hline \hline
Accomodation ID&87\\
Hotel Name&\begin{CJK}{UTF8}{ipxm}ベルビューガーデンホテル関西空港
\end{CJK} (Bellevue Garden Hotel Kansai Airport)\\
\hline

\end{tabular}
\centering
\\(b) Hotel Master Data
 \caption{\label{tb:corpus} Corpus Specification of Rakuten Travel Reviews Data (English translation is given in parentheses)}
\end{table*}

\section{Related Work}
One of the major review datasets was released by Amazon, Inc. back in 2020~\cite{keung:20}. 
As an initiative separate from the Amazon, \newcite{hou:24}~released up-to-date Amazon review datasets named ``AMAZON REVIEWS 2023'', ranging from May 1996 to September 2023~\footnote{\url{https://amazon-reviews-2023.github.io}}.
There are user-generated review datasets in the hotel domain, mainly crawled from leading online travel platforms such as TripAdvisor and Booking.com. For instance, \newcite{Fang:16} crawled 41k reviews posted in 2014 for attractions in New Orleans to analyze the perceived helpfulness of reviews. \newcite{Tsamis:21} crawled 65k English hotel reviews from TripAdvisor and trained a DNN on helpfulness prediction. The Kaggle dataset~\footnote{\url{https://www.kaggle.com/datasets/jiashenliu/515k-hotel-reviews-data-in-europe/discussion?sort=undefined}} contains 515k reviews crawled from Booking.com published between August 2015 and August 2017.
Finally, \newcite{Igebaria:24} created a corpus of English reviews posted in 2023 for predicting individual review helpfulness.
Although these datasets are large-scale, they do not provide sufficiently long period, covering three significant eras in the Travel industry: pre-COVID, mid-COVID, and post-COVID.
None of these studies focuses on releasing review texts for more than 15 years in the hotel domain.
\section{Rakuten Travel Dataset}
\subsection{Corpus Specification}
Our corpus contains ``User Evaluation Data'' and ``Hotel Master Data'', as described in~Table\ref{tb:corpus}.
Unlike previous Rakuten Travel Data, our updated corpus includes reviews data for the past five years (i.e., 2020 to 2024). 
User evaluation data consists of the following 18 columns.
\paragraph{Reviewer ID} Reviewer ID represents a masked user name that starts with prefix``user\_''. Our corpus has 2,505,326 distinct users.
\paragraph{Review Date} Review date refers to the posted date of review, formatted as YYYY-MM-DD HH:MM:SS.
\paragraph{Accommodation ID} This column represents the target hotel reviewed. We focus on hotels located in Japan, operating as of July 2, 2025. Our corpus has 27,603 distinct hotels.
\paragraph{Plan ID}
In the context of hotel reservations, a ``plan'' refers to a campaign offered by accommodation, combining staying with associated services and benefits. It goes beyond simply ``booking a room'', and is sold with various elements incorporated to meet the diverse needs and purposes of guests. There are 985,678 unique plan IDs in our corpus. 
\paragraph{Plan Title}
Plan title represents campaign contents expressed in a short sentence.
\paragraph{Room Type} Room type represents a pre-defined category name for a room. Our corpus has 103,256 room types. 
\paragraph{Room Name} Room name is associated with room features such as smoking availability, room size and capacity, expressed in a short sentence.
\paragraph{Purpose}
This column refers to stay purpose for a reviewer which can be selected from one of the three types:\\\begin{CJK}{UTF8}{ipxm}レジャー\end{CJK} (Leisure), \begin{CJK}{UTF8}{ipxm}出張\end{CJK} (Business), \\and \begin{CJK}{UTF8}{ipxm}その他\end{CJK} (others).
\paragraph{Accompanying Group}
This column refers to who accompanies a reviewer. Reviewer can select one of the following options: \begin{CJK}{UTF8}{ipxm}一人\end{CJK} (solo),\\ \begin{CJK}{UTF8}{ipxm}家族\end{CJK} (family),\begin{CJK}{UTF8}{ipxm}恋人\end{CJK} (couple), \begin{CJK}{UTF8}{ipxm}友達\end{CJK} (friend), \\\begin{CJK}{UTF8}{ipxm}仕事仲間\end{CJK} (co-worker), and \begin{CJK}{UTF8}{ipxm}その他\end{CJK} (others).
\paragraph{User Rating}
Rating was posted by a reviewer on a five-point scale across pre-defined six aspect categories, namely, Location, Room, Food, Bath, Services, and Amenity, as well as an overall rating. In total, our corpus has seven columns for rating scores. The value ``5'' indicates the highest evaluation, whereas a rating ``1'' refers to the lowest evaluation.
The value ``0'' indicates that the user was unable to evaluate the corresponding aspect category since there was no opportunity to experience it during the stay.
\paragraph{Review} Review has a character limit of less than 1,000 characters.
\paragraph{Reply} This column refers to a response from accommodation staff to the review.
\begin{figure*}[ht]
    \begin{center}
   \includegraphics[width=10cm]{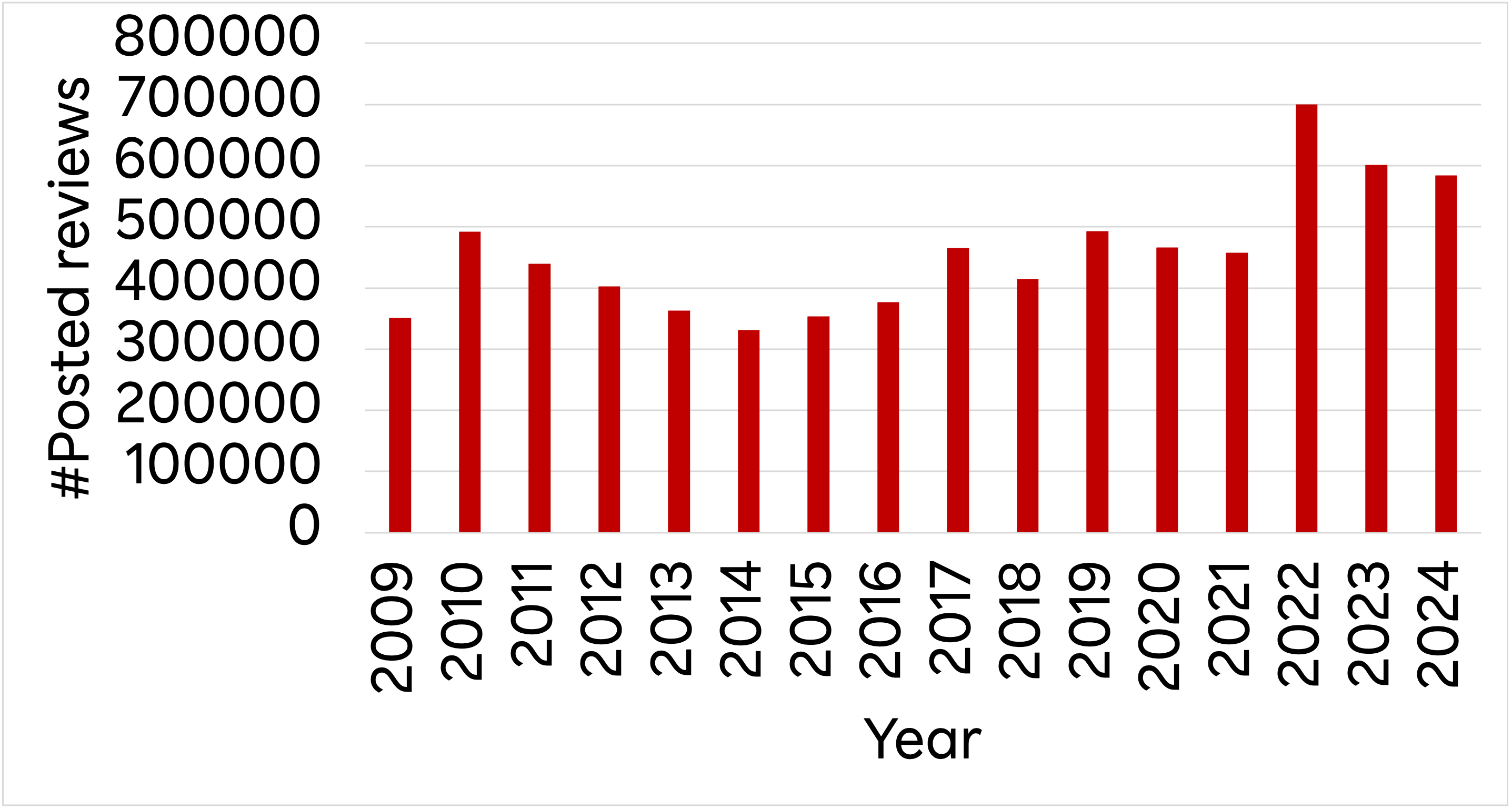}
    \caption{Number of posted reviews by year}
    \label{fig:review_num}
    \end{center}
\end{figure*}
\subsection{Data Statistics}

\begin{figure*}[ht]
    \begin{center}
   \includegraphics[width=13cm]{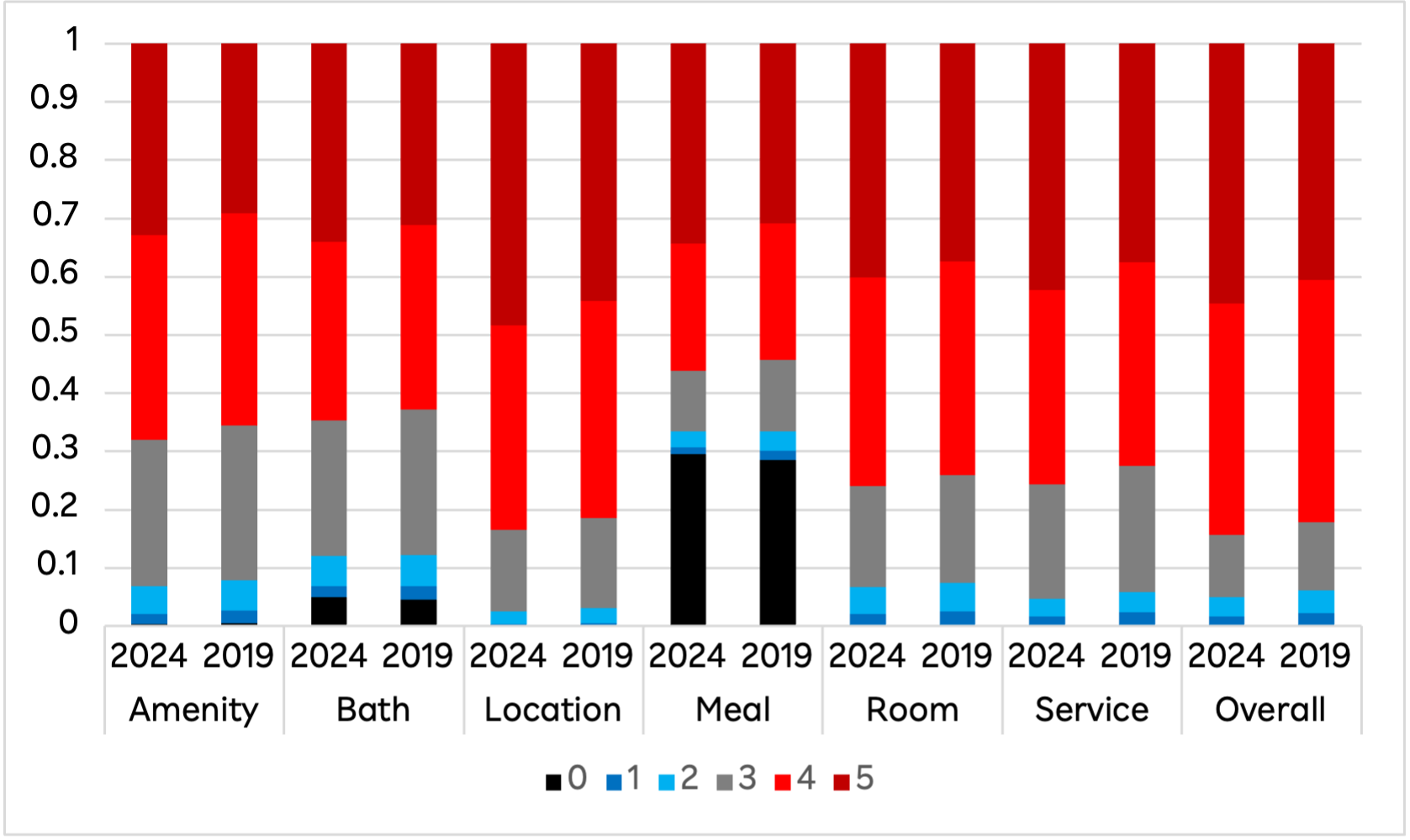}
    \caption{Rating Distribution on an aspect-by-aspect basis for 2019 and 2024}
    \label{fig:rating_dist}
    \end{center}
\end{figure*}

Figure~\ref{fig:review_num} shows the number of posted reviews by year in our corpus. In the most recent three years following the COVID-19 pandemic, the number of reviews posted has increased, which implies a growing interest in and demand for travel. Figure~\ref{fig:rating_dist} shows rating distribution by each aspect for 2019 and 2024. 

To verify the data drift between old and new data, we observe a gap in word distribution in reviews between 2019 and 2024, as evidenced by the chi-square test of words.
We first divided a review into one or more sentences with the heuristic rules. Then, we tokenized review sentences with MeCab and the MeCab-ipadic-neologd dictionary, and counted the frequencies of unigrams, bigrams, and trigrams. We found 4,453,580 tokens with statistically significant differences at 1\% level.
Let us discuss factors causing data drift, especially focusing on the following factors, as indicated by the higher odds rate and chi-square value in Table~\ref{tb:dist}.

\section{Factors Causing Data Drift}
\begin{table*}[ht]
\begin{center}
 \begin{tabular}{lrrr}
 \hline
 \multicolumn{1}{c}{\multirow{2}{*}{Token}}&\multicolumn{2}{c}{Document Frequency}&Odds rate\\
 \cline{2-3}
 &2024&2019&\\
 \hline
 \begin{CJK}{UTF8}{ipxm}コロナ (Corona)\end{CJK}&1152&20&48.79\\
 \begin{CJK}{UTF8}{ipxm}コロナ禍 (COVID-19 pandemic)\end{CJK}&530&0&$\infty$\\
 \begin{CJK}{UTF8}{ipxm}リモートワーク\end{CJK} (remote work)&99&2&41.78\\
 \begin{CJK}{UTF8}{ipxm}テレワーク\end{CJK} (telework)&85&1&71.74\\
 \begin{CJK}{UTF8}{ipxm}ワーケーション\end{CJK} (workcation)&120&0&$\infty$\\

 \begin{CJK}{UTF8}{ipxm}非接触\end{CJK} (contactless) & 115 & 0&$\infty$\\
 \begin{CJK}{UTF8}{ipxm}QRコード\end{CJK} (QR code)&819&20&31.50\\ 
 \begin{CJK}{UTF8}{ipxm}モバイル-オーダー\end{CJK} (mobile order)&55&0&$\infty$\\
 
 \begin{CJK}{UTF8}{ipxm}ホカンス (staycation)\end{CJK}&68&0&$\infty$\\
  \begin{CJK}{UTF8}{ipxm}ウェルカムバー (welcome bar)\end{CJK}&242&0&$\infty$\\
 ReFa&900&0&$\infty$\\
 \begin{CJK}{UTF8}{ipxm}ナノバブル\end{CJK} (nanobubble)&29&0&$\infty$\\
 \begin{CJK}{UTF8}{ipxm}ドームテント (dome tent)\end{CJK}&128&0&$\infty$\\
 
 \begin{CJK}{UTF8}{ipxm}フードロス\end{CJK} (food waste)&85&6&11.96\\
  SDGs & 253 & 1&196.73\\
  \begin{CJK}{UTF8}{ipxm}環境-配慮\end{CJK} (environmentally-friendly)&151&38&3.35\\
  \begin{CJK}{UTF8}{ipxm}アメニティ-環境-配慮\end{CJK} (amenity-environmentally-friendly&22&0&$\infty$\\
 \begin{CJK}{UTF8}{ipxm}エモい\end{CJK}(emotional)&23&0&$\infty$\\
 \begin{CJK}{UTF8}{ipxm}ととのう\end{CJK} (sauna trance)&177&15&9.96\\
 \begin{CJK}{UTF8}{ipxm}インバウンド\end{CJK} (inbound)& 1337 & 263&4.30\\
 \begin{CJK}{UTF8}{ipxm}高騰\end{CJK} (soar) & 1286 & 178 & 6.12\\
 \begin{CJK}{UTF8}{ipxm}物価-高騰\end{CJK} (soar)&126&3&35.45\\
 \begin{CJK}{UTF8}{ipxm}物価高\end{CJK} (soar) & 249 & 1 & 210.28\\
  
 \hline
 
 \end{tabular}
 \caption{\label{tb:dist}Document frequency of words with statistically significant difference between 2019 and 2024 (p<0.01)}
 \end{center}
\end{table*}
\paragraph{COVID-19}
Since the advent of the COVID-19 pandemic in 2020, the term ``\begin{CJK}{UTF8}{ipxm}コロナ\end{CJK}  (Corona)'' has frequently appeared in review texts, consistently referring to the coronavirus whereas this word has different meaning of hotspring and facility in 2019 such as ``\begin{CJK}{UTF8}{ipxm}コロナの湯\end{CJK} (corona-hotspring)'' and ``\begin{CJK}{UTF8}{ipxm}コロナワールド (corona-world)\end{CJK}''.
Reflecting on changing of working style due to COVID-19, the words associated with ``work from home'' were used in 2024 such as ``\begin{CJK}{UTF8}{ipxm}ワーケーション\end{CJK}'' and '\begin{CJK}{UTF8}{ipxm}リモート\end{CJK}''.
The widespread adoption of services designed with social distancing in mind has led to a frequent increase in the appearance of terms associated with non-contact interactions services such as `\begin{CJK}{UTF8}{ipxm}非接触 \end{CJK}(contactless)'', ``\begin{CJK}{UTF8}{ipxm}QRコード\end{CJK} (QR code)'', and ``\begin{CJK}{UTF8}{ipxm}モバイルオーダー\end{CJK} (mobile order)'', 
The COVID-19 pandemic has led to restrictions on eating out. As a result, there has been a surge in hotels offering accommodations where guests can enjoy food and drinks on-site, fuelled by the rise in stay-at-home demand. 

The trend of refraining from long-distance travel due to the COVID-19 pandemic significantly influenced the rapid spread of ``hocance'', a portmanteau combining ``hotel'' and ``vacance'' that refers to a style emphasizing fully enjoying hotels' services. A model trained on past years' data cannot deal with which aspect categories are mentioned in the sentence ``\begin{CJK}{UTF8}{ipxm}ホカンス大成功でした！\end{CJK}''. 

\paragraph{SDGs}
The trigram words ``\begin{CJK}{UTF8}{ipxm}環境-配慮-する\end{CJK} (environmentally friendly)'' and ``\begin{CJK}{UTF8}{ipxm}アメニティ-環境-配慮\end{CJK} (get amenity eco-friendly)'' are much more frequently in the review texts published in 2024 compared with 2019.
The increasing prominence of the SDGs within the hotel industry suggests that customers are now evaluating and providing feedback on environmentally conscious elements, particularly practices such as amenity distribution.
\paragraph{New Features and New Trends}
To enhance the user experience, the hotel industry often introduces new features. For instance, ``ReFa’' and \begin{CJK}{UTF8}{ipxm}``ナノバブル''\end{CJK} frequently appeared in the review texts in 2024 by making the quality of shower equipment better, whereas these words were not included in 2019 at all since the nationwide sale of ReFa began in 2020. When we assume tackling the aspect-category detection task, the sentence ``\begin{CJK}{UTF8}{ipxm}ReFaシリーズも試せてよかった！''\end{CJK} is likely to be difficult to classify as equipment.


\paragraph{Inbound Tourism}
Compared to 2019, the term "inbound" is now referenced significantly more often. This surge in usage can be attributed to the rise of social media-popular ``Instagrammable'' spots as new tourist attractions, coupled with a significant increase in foreign visitors, spurred by a favorable exchange rate for the yen and inexpensive living costs from Western countries' perspective.
\paragraph{Inflation and Price Hikes}
\begin{CJK}{UTF8}{ipxm}``物価-高騰 (soar)''\end{CJK} and \begin{CJK}{UTF8}{ipxm}``物価高 (soar)''\end{CJK} frequently appear in 2024, whereas the phrase was hardly ever mentioned in 2019 since soaring hotel prices are critical issue of Japan economy.
\paragraph{Slang Words}

``\begin{CJK}{UTF8}{ipxm}エモい\end{CJK}'' is originated from ``emotional''.
This is a very convenient and ambiguous term that encompasses a diverse range of emotions and sensations such as "feeling deeply moved," ``resonating deeply,``nostalgic,``poignant, and ``having a unique atmosphere. 
Additionally, although there was no significant difference, other new slang words appeared in 2024 whereas in 2019 these did not such as ``\begin{CJK}{UTF8}{ipxm}チルい\end{CJK} derived from ``Chilling out'' and ``\begin{CJK}{UTF8}{ipxm}沼る\end{CJK} (obsessed)''.
Such new slang words make NLP tasks, such as sentiment analysis, more difficult.
\section{Conclusion}
This paper presents an up-to-date Rakuten Travel reviews corpus to facilitate researchers at academic institutions, particularly universities laboratories, in conducting research with real-world data. To the best of our knowledge, there have been no prior attempts to publicly release hotel review data with a collection period exceeding 15 years. Through statistical analysis of word distributions between 2019 and 2024, we identified key factors causing data drift in the hotel review domain. The COVID-19 pandemic fundamentally transformed the hospitality industry by introducing new vocabulary related to remote work, contactless services, and hygiene management. With the increasing importance of SDGs, customers are increasingly evaluating and providing feedback on environmentally conscious initiatives, particularly regarding amenity distribution practices. Furthermore, a surge in inbound tourism, driven by social media trends and favorable exchange rates, has significantly impacted review content. These findings underscore the importance of using up-to-date data for model training and evaluation in real-world business applications. 
\section*{Limitations}
We grant academic research institute the access of our corpus for academic research purposes.




\bibliography{anthology,skeiji}
\bibliographystyle{acl_natbib}







\end{document}